\documentclass{article}

\usepackage{arxiv}

\usepackage{amsfonts}  
\usepackage{amsmath} 
\usepackage{float}
\usepackage{longtable}
\usepackage{array}
\usepackage{makecell}
\usepackage{siunitx}
\usepackage[T1]{fontenc}    
\usepackage{hyperref}       
\usepackage{url}            
\usepackage{hyperref}
\hypersetup{
    breaklinks=true,
    colorlinks=true,
    linkcolor=blue,
    urlcolor=blue,
    citecolor=blue
}

\usepackage{booktabs}       
\usepackage{amsfonts}       
\usepackage{nicefrac}       
\usepackage{microtype}      
\usepackage{cleveref}       
\usepackage{lipsum}         
\usepackage{graphicx}
\usepackage{natbib}
\usepackage{doi}
\usepackage{parskip} 

\title{Adaptive Stiffness: A Biomimetic Robotic System with Tensegrity-Based Compliant Mechanism}

\date{}

\newif\ifuniqueAffiliation
\uniqueAffiliationtrue

\ifuniqueAffiliation 
\author{ \href{https://orcid.org/0009-0001-2682-138X}{\includegraphics[scale=0.06]{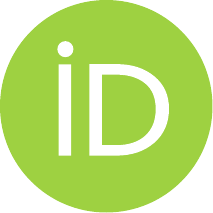}\hspace{1mm}Po-Yu Hsieh}\thanks{The scripts used in this study can be found on GitHub at the following repository: \url{https://github.com/poyuhs/model-free-robotics.git}} \\
	Graduate Institute of Architecture\\
	National Yang Ming Chiao Tung University\\
	Hsinchu, Taiwan \\
	\texttt{kevinhsieh870118@arch.nycu.edu.tw} \\
	\And
	\href{https://orcid.org/0000-0002-8362-7719}{\includegraphics[scale=0.06]{orcid.pdf}\hspace{1mm}June-Hao Hou} \\
	Graduate Institute of Architecture\\
	National Yang Ming Chiao Tung University\\
	Hsinchu, Taiwan \\
	\texttt{jhou@arch.nycu.edu.tw} \\
}
\else
\usepackage{authblk}

\newbox{\orcid}\sbox{\orcid}{\includegraphics[scale=0.06]{orcid.pdf}} 
\author[1]{%
	\href{https://orcid.org/0000-0000-0000-0000}{\usebox{\orcid}\hspace{1mm}David S.~Hippocampus\thanks{\texttt{hippo@cs.cranberry-lemon.edu}}}%
}
\author[1,2]{%
	\href{https://orcid.org/0000-0000-0000-0000}{\usebox{\orcid}\hspace{1mm}Elias D.~Striatum\thanks{\texttt{stariate@ee.mount-sheikh.edu}}}%
}
\affil[1]{Department of Computer Science, Cranberry-Lemon University, Pittsburgh, PA 15213}
\affil[2]{Department of Electrical Engineering, Mount-Sheikh University, Santa Narimana, Levand}
\fi

\renewcommand{\headeright}{}
\renewcommand{\undertitle}{}
\renewcommand{\shorttitle}{Adaptive Stiffness: A Biomimetic Robotic System with Tensegrity-Based Compliant Mechanism}

\hypersetup{
pdftitle={Adaptive Stiffness: A Biomimetic Robotic System with Tensegrity-Based Compliant Mechanism},
pdfsubject={cs.RO},
pdfauthor={Po-Yu Hsieh, June-Hao Hou},
pdfkeywords={Biomimetic, Adaptive Robot, Tensegrity, Compliant Mechanism, Rigid-Flex Coupling},
}

\begin{document}
\maketitle

\begin{abstract}
	Biomimicry has played a pivotal role in robotics. In contrast to rigid robots, bio-inspired robots exhibit an inherent compliance, facilitating versatile movements and operations in constrained spaces. The robot implementation in fabrication, however, has posed technical challenges and mechanical complexity, thereby underscoring a noticeable gap between research and practice. To address the limitation, the research draws inspiration from the unique musculoskeletal feature of vertebrate physiology, which displays significant capabilities for sophisticated locomotion. The research converts the biological paradigm into a tensegrity-based robotic system, which is formed by the design of rigid-flex coupling and a compliant mechanism. This integrated technique enables the robot to achieve a wide range of motions with variable stiffness and adaptability, holding great potential for advanced performance in ill-defined environments. In summation, the research aims to provide a robust foundation for tensegrity-based biomimetic robots in practice, enhancing the feasibility of undertaking intricate robotic constructions.
\end{abstract}


\section{Introduction}
In recent decades, digital fabrication has relied on robotics for mechanical precision and technological efficiency. The traditional use of rigid robots has been exploited in various applications, especially manufacturing, due to their competence in dealing with massive and repetitive workloads in fixed spaces \citep{trivedi2008soft}. Nevertheless, their inability to conform to external impacts or environmental changes is evident. As more advanced demands, such as operation and exploration in extreme terrains \citep{kobayashi2022soft}, gradually increased, the inspiration from nature has led to the emergence of biomimicry in robotics. Developers across disciplines, including biology, engineering, and architecture, seek solutions to replicate the adaptability and mobility observed in biological systems.
\par Unlike conventional rigid robots, most biological structures rely on a harmonious blend of rigid and soft components. In vertebrates, the appendages are based on a musculoskeletal system composed of bones, muscles, tendons, and joints. The integration of rigid and soft elements constitutes a dynamic form, displaying versatile mobility and high compliance in degrees-of-freedom. Rigid bones undergo compression and comprise the framework of the entire structure. The muscles, on the contrary, maintain the stability of the system via a tensional network. This rigid-flex connection between bones and muscles, driven by tendons as actuators, is a crucial feature of this bio-inspired system.
\par In terms of biomechanics, the tendon-driven actuation allows precise, centralized control over bone movements through a continuous force distribution. Once the structure is pulled by tension from the tendons, the stress will be distributed through the tensional network. The configuration therefore changes to match the new equilibrated form. In comparison with the rigid counterpart, this bio-inspired mechanism excels in driving efficiency and range of motions, offering a paradigm shift in robotic design.
\par To address the existing constraints of rigid robots and explore the implementation of biomimetic robotics, this research draws inspiration from biomechanics, specifically vertebrate physiology \citep{zappetti2020variable}. The structural composition allows vertebrates to possess inherent mechanical advantages, including versatile movement and effective adaptability. This biological system relies on an integrated framework between rigid and soft bodies, which can be characterized by a discrete set of compressed components and a continuum of tensional network. The structural amalgamation provides high compliance, flexibility, driving efficiency, and effective force distribution with a self-balancing mechanism \citep{liu2022review}. Moreover, depending on their current condition or intention, vertebrates can switch the structural stiffness to change their configuration, which enhances the environmental adaptability. Research has confirmed that this biological system aligns with the tensegrity structure, providing an effective basis for robotic bodies \citep{lessard2016bio}. Similar to vertebrate appendages, this unique structure is also comprised of rigid and soft components \citep{motro2003tensegrity} with dexterous reconfigurability.
\par Building upon this anatomical and mechanical concept, the research has developed a hybrid robotic system based on a numerical form of tensegrity structure. This proposed structure is derived from a linear augmentation of prismatic tensegrity units \citep{zhang2015tensegrity}, imitating the spine-like attributes commonly seen in vertebrates. Furthermore, the research presents comprehensive information about the robot implementation, grafting the tensegrity-based compliant mechanism onto a mechatronic control system with programmable tendon-driven actuators. This innovation allows the robot to perform variable motions, demonstrating a feasible framework for tensegrity-based robots in architecture, engineering, and construction (AEC) industries.

\begin{figure}[H]
	\begin{center}
		\includegraphics[width=\textwidth]{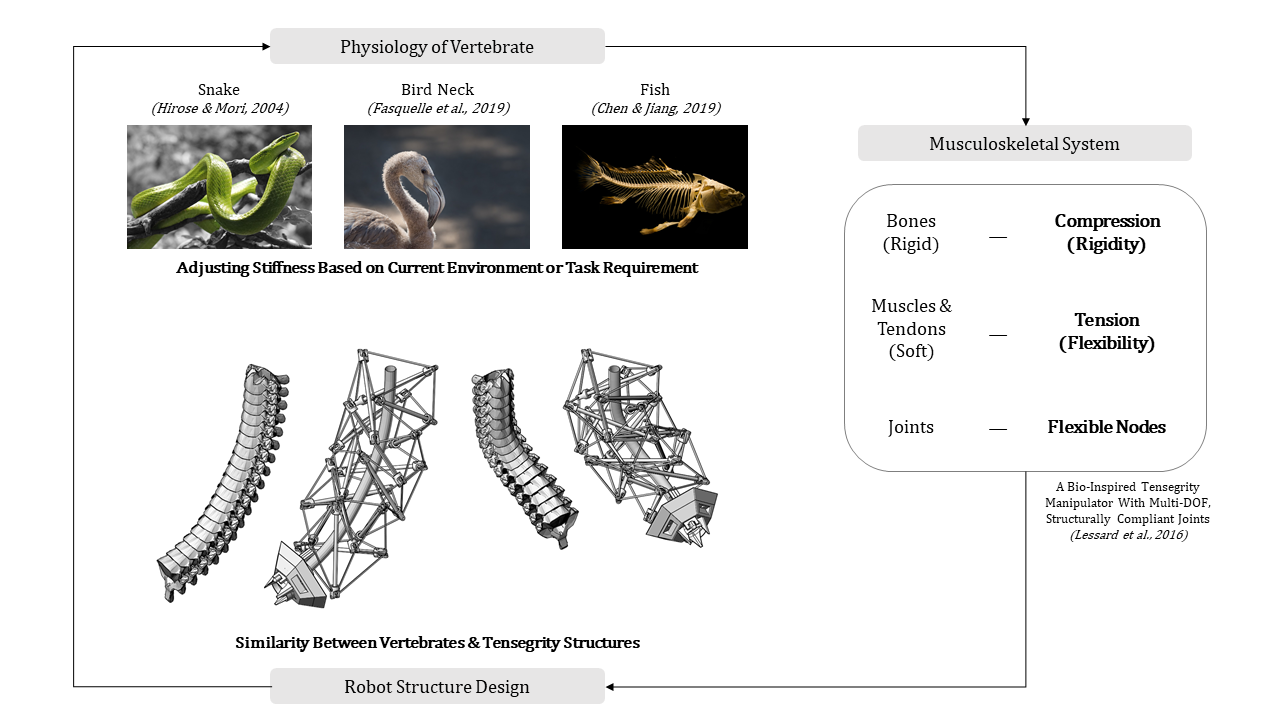}
		\caption{Design methodology.}\label{fig_methodology}
	\end{center}
\end{figure}

\section{Related Works}
Biomimetic robots based on tensegrity structures are constructed by mimicking the geometric form, function, and kinematic principles found in the nature world. Researchers in the field of architecture and mechanical engineering have conducted scientific projects attempting to graft the structural properties onto robotics. The principal goal of these bio-inspired robots is to represent the structure and behavior of specific living organism or appendages, such as fish \citep{chen2019swimming}, human shoulder \citep{li2022robotic}, bird neck \citep{fasquelle2019dynamic}, and snakes \citep{hirose2004biologically}, broadening the application of biomimetic robots in practice. Research has indicated that tensegrity structures can be manipulated by either changing the length of compressive components or pulling on certain segment of tensional members (Hanaor and Levy, 2001). The former method provides a deployable capability, while the latter allows the robot to achieve variable movements. To further explore the mobility and optimize the driving efficiency, the research adopts the pulling approach and converts it into the tensegrity-based compliant mechanism.

\section{Robot Implementation}

\subsection{Deformable Tensegrity}

\subsection*{Elastic nature}
\begin{figure}[hbp]
	\begin{center}
		\includegraphics[width=.85\textwidth]{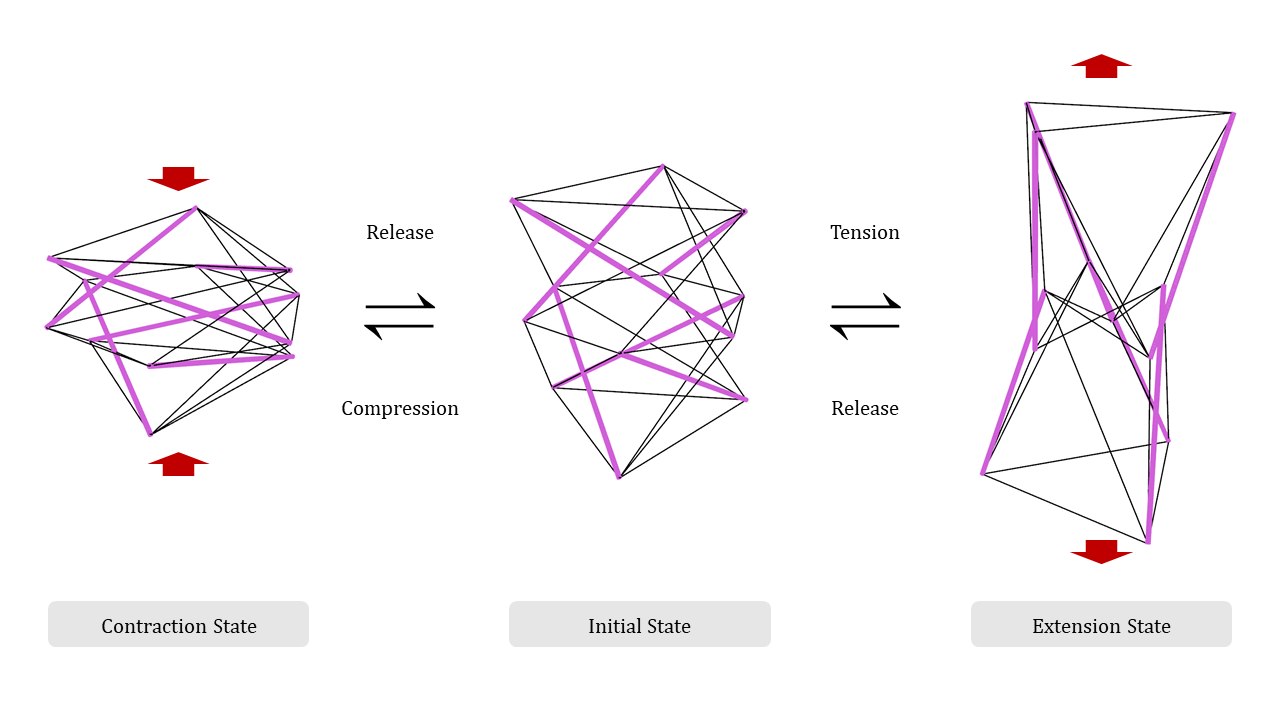}
		\caption{Two types of deformation.}\label{fig_deformation}
	\end{center}
\end{figure}

The design starts with the the deformation behavior of a tensegrity-based robot body through three distinct states: initial state, contraction, and extension.Each state is visually represented by a geometric network of interconnected cables and struts (Figure~\ref{fig_deformation}), with the black lines indicating the cables (tensile components) and the purple lines indicating the struts (compressive components). 
The transitions between these states involve the release and application of compressive and tensile forces. When moving from the contraction state to the initial State, the release of compression allows the structure to return to its equilibrium configuration. Similarly, transitioning from the initial state to the extension state involves the application of tensile forces, leading to the elongation of the structure. The reversibility of these transitions highlights the elastic nature of the tensegrity-based robot body, capable of deforming and returning to its original shape under varying external forces.
\par The initial state represents the equilibrium condition of the robot body, where no external forces are applied, and the system is at rest. In the contraction state, the robot body is subjected to compressive forces, causing it to compact and shorten along the vertical axis. This dense configuration provides stiffness compared to the initial state. The extension state occurs when the robot body is subjected to tensile forces, forcing it to stretch along the vertical axis. This elongated configuration, on the other hand, offers noticeable flexibility.

\vspace{-.4cm}
\vspace{.4cm}

\subsection{Parametric Modeling}
\par
\begin{figure}[H]
	\begin{center}
		\includegraphics[width=.7\textwidth]{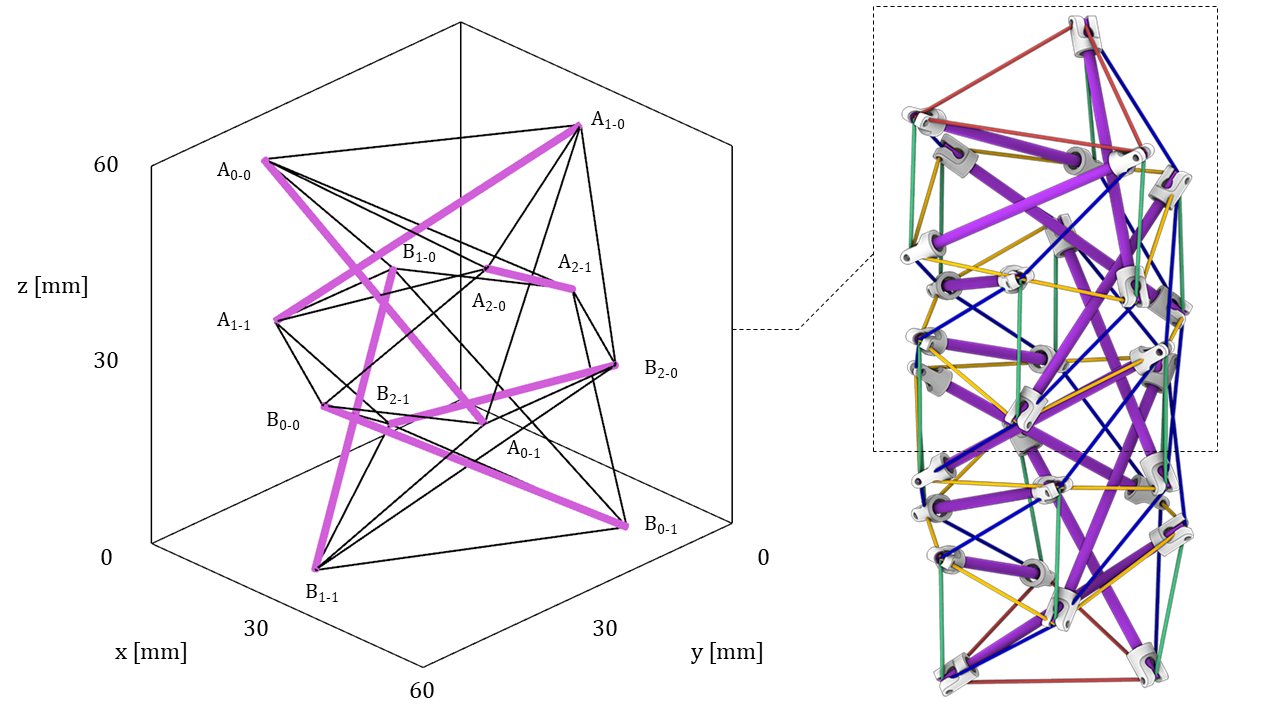}
		\caption{Left: Tensegrity vertices. Right: Linear augmentation.}\label{fig_vertices}. 
	\end{center}
\end{figure}

\begin{figure}[H]
	\begin{center}
		\includegraphics[width=.8\textwidth]{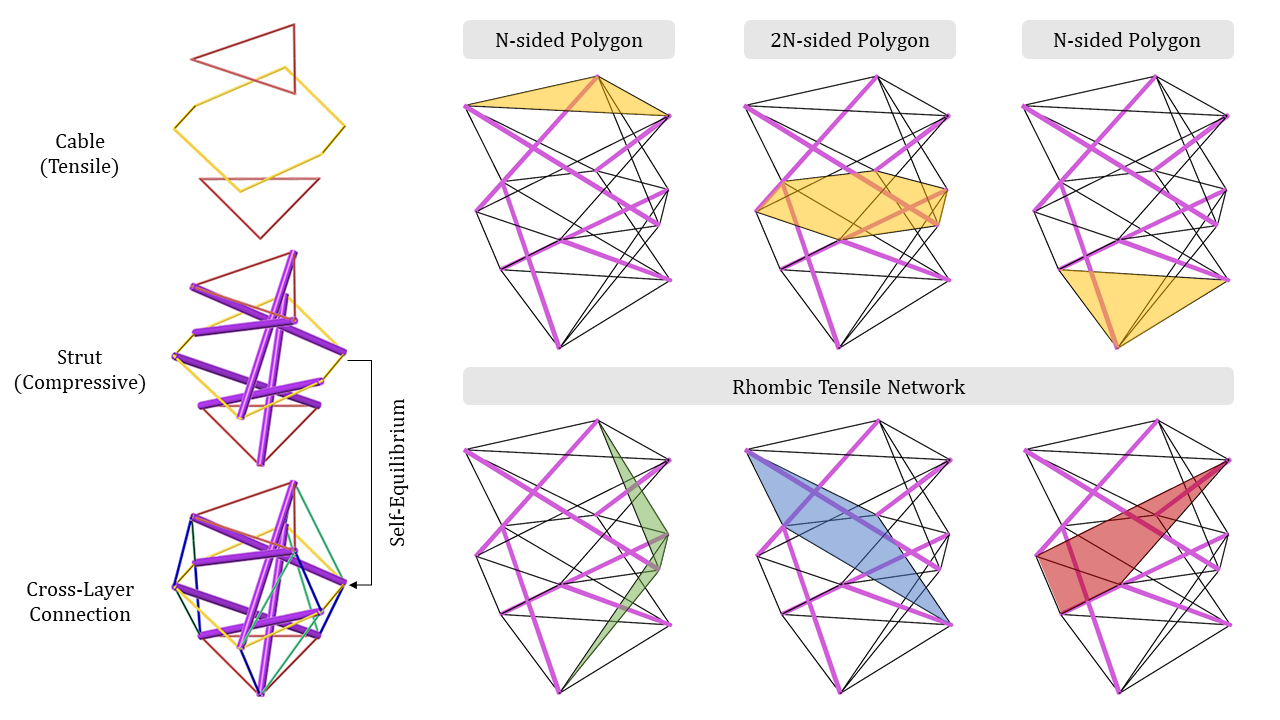}
		\caption{Structural composition.}\label{fig_parametricModeling}
	\end{center}
\end{figure}

 To imitate the biological system based on tensegrity structures, the research employs the adaptive force density approach to deal with the form-finding issues. As presented in Table~\ref{table_parametrized} and Figure~\ref{fig_vertices} (left), this multi-layer form is comprised of a tensional cable network and a discontinuous set of struts. The tensional network includes four types of cables: horizontal, vertical, diagonal, and saddle cable. The horizontal cables form a \(n\)-sided polygon on both ends, and the saddle cables create a 2\(n\)-sided polygon in the middle (\( n \in \mathbb{N}_{\text{odd}} \), \( n \geq 3 \)). The vertical and diagonal cables establish the interlayer links. The number of layers (\(m\)) can be considered as (3\(p\)+3), where \( p \in \mathbb{N} \). This robot structure is comprised of triangular plates on both ends, hexagonal plates in the middle, and a rhombic network in a circumferential pattern (Figure~\ref{fig_parametricModeling}). The proposed numerical robot model is adjustable in dimensions and composition with parametric design process. This provides a customizable foundation and reduces the mechanical complexity of biomimetic robots.
\par The design starts with the prismatic tensegrity. Through the application of external forces, the structure reconfigures with flexibility, displaying its inherent compliance. By augmenting the form with multiple tensegrity units, a spine-like topology is generated and its deformability is amplified as demonstrated in Figure~\ref{fig_vertices} (right). 

\renewcommand{\arraystretch}{1.5}
\begin{center}
    \begin{longtable}{l
        p{3cm}
        p{3cm}
        p{3cm}
        p{3cm}}
        \caption{Parametric modeling (the 'order' can be referred to Figure~\ref{fig_vertices}).} \\
        \toprule
        \textbf{} & \textbf{Horizontal Cables} & \textbf{Saddle Cables} & \textbf{Vertical Cables} & \textbf{Diagonal Cables}\\
        \midrule
        \endfirsthead
        
        \caption[]{(continued)} \\
        \toprule
        \textbf{} & \textbf{Horizontal Cables} & \textbf{Saddle Cables} & \textbf{Vertical Cables} & \textbf{Diagonal Cables} \\
        \midrule
        \endhead
        
        \midrule
        \multicolumn{5}{r}{\textit{continued on next page}} \\
        \endfoot
        
        \bottomrule
        \endlastfoot
        
        \textbf{Position} & Top, Bottom & Middle & Cross-layer & Cross-layer \\
        \textbf{Order} & 
        \makecell[l]{\( A_{0-0} \)----\( A_{1-0} \) \\ \( A_{1-0} \)----\( A_{2-0} \) \\ \( A_{2-0} \)----\( A_{0-0} \) \\ \( B_{0-1} \)----\( B_{1-1} \) \\ \( B_{1-1} \)----\( B_{2-1} \) \\ \( B_{2-1} \)----\( B_{0-1} \)} & 
        \makecell[l]{\( A_{0-1} \)----\( B_{0-0} \) \\ \( B_{0-0} \)----\( A_{1-1} \) \\ \( A_{1-1} \)----\( B_{1-0} \) \\ \( B_{1-0} \)----\( A_{2-1} \) \\ \( A_{2-1} \)----\( B_{2-0} \) \\ \( B_{2-0} \)----\( A_{0-1} \)} & 
        \makecell[l]{\( A_{0-0} \)----\( A_{2-1} \) \\ \( A_{1-0} \)----\( A_{0-1} \) \\ \( A_{2-0} \)----\( A_{1-1} \) \\ \( B_{0-0} \)----\( B_{2-1} \) \\ \( B_{1-0} \)----\( B_{0-1} \) \\ \( B_{2-0} \)----\( B_{1-1} \)} & 
        \makecell[l]{\( A_{0-0} \)----\( B_{0-0} \) \\ \( A_{1-0} \)----\( B_{1-0} \) \\ \( A_{2-0} \)----\( B_{2-0} \) \\ \( A_{0-1} \)----\( B_{0-1} \) \\ \( A_{1-1} \)----\( B_{1-1} \) \\ \( A_{2-1} \)----\( B_{2-1} \)} \\ 
        \textbf{Amount} & \(h\)=2\(n\) & \(s\)=2\(n\)(\(m\)-2) & \(v\)=\(n\)(\(m\)-1) & \(d\)=\(n\)(\(m\)-1)  \\
        \textbf{\(m\)-Layer} & \multicolumn{4}{c}{\(m\) = 3, 6, 9, \ldots, 3\(p\)+3, where \( p \in \mathbb{N} \)} \\ 
		\vspace{-.5cm}

        \label{table_parametrized}
        
    \end{longtable}
\end{center}
\vspace{-1cm}
	
\section{Mechanical Design}

\subsection{Rigid-Flex Coupling and Compliant Mechanism}
\begin{figure}[H]
	\begin{center}
		\includegraphics[width=.75\textwidth]{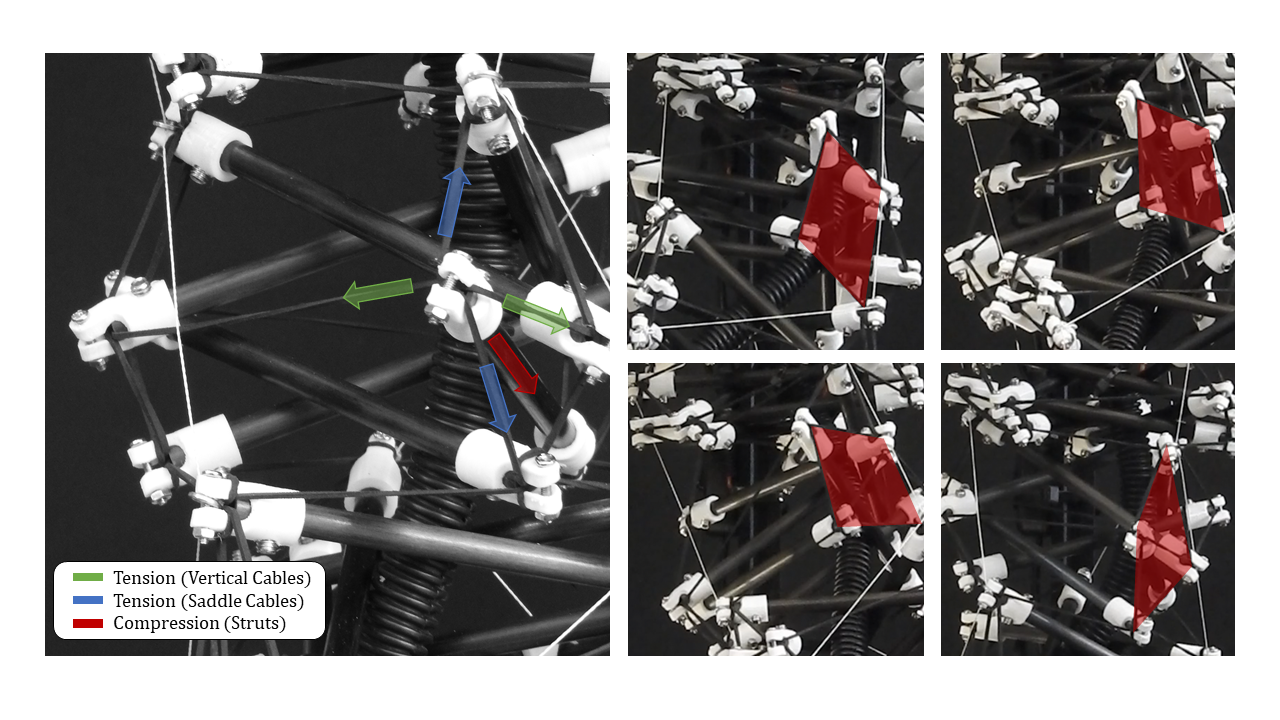}
		\caption{Left: Rigid-Flex coupling. Right: Rhombic configuration transformations.}\label{fig_rigidFlexCoupling}
	\end{center}
\end{figure}
\begin{figure}[H]
	\begin{center}
		\includegraphics[width=.7\textwidth]{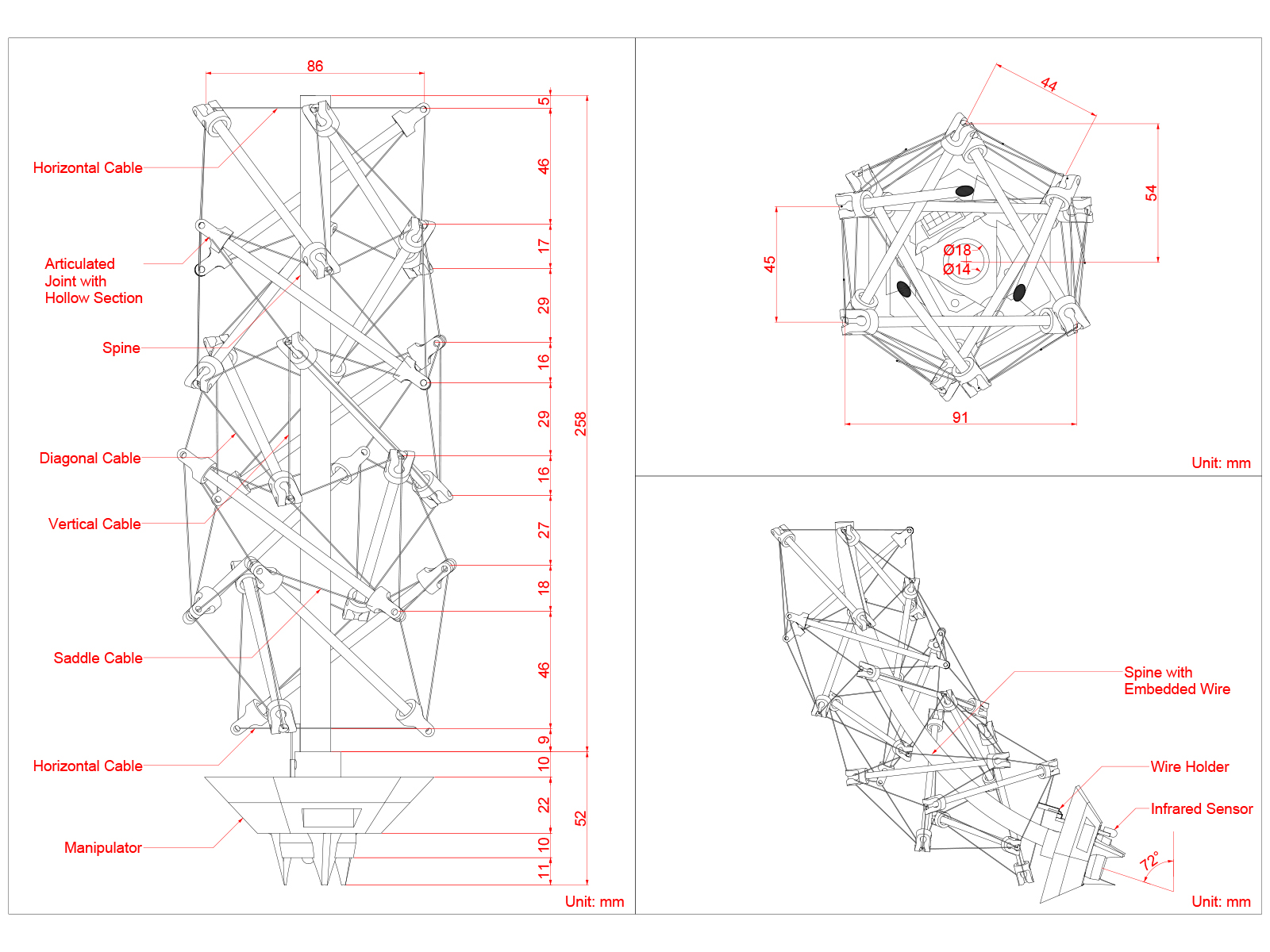}
		\caption{CAD drawings (left: right view. top-right: top view. bottom-right: equipments).}\label{fig_cad}
	\end{center}
\end{figure}

The efficacy of the mechanical system has a profound influence on robotic performance, tensegrity-based robots in particular. To achieve versatile mobility and precise manipulation, the self-equilibration based on tensegrity joints needs to be effectively carried out. Thus, the research has developed a rigid-flex coupling, transforming the entire robot body into a dynamic form with compliant mechanism. Combining the mechanical advantages of both rigid and soft structures, the coupling ensures effective force distributions through the rhombic tensile network (Figure~\ref{fig_rigidFlexCoupling}) . The coupling also ensures adaptive changes in the structural stiffness, facilitating advanced movements or load-bearing operations. By applying the coupling to the struts and cables, the entire robot body is constructed (Figure~\ref{fig_cad}).

\subsection{Mechatronic Control System}

\begin{figure}[H]
	\begin{center}
		\includegraphics[width=\textwidth]{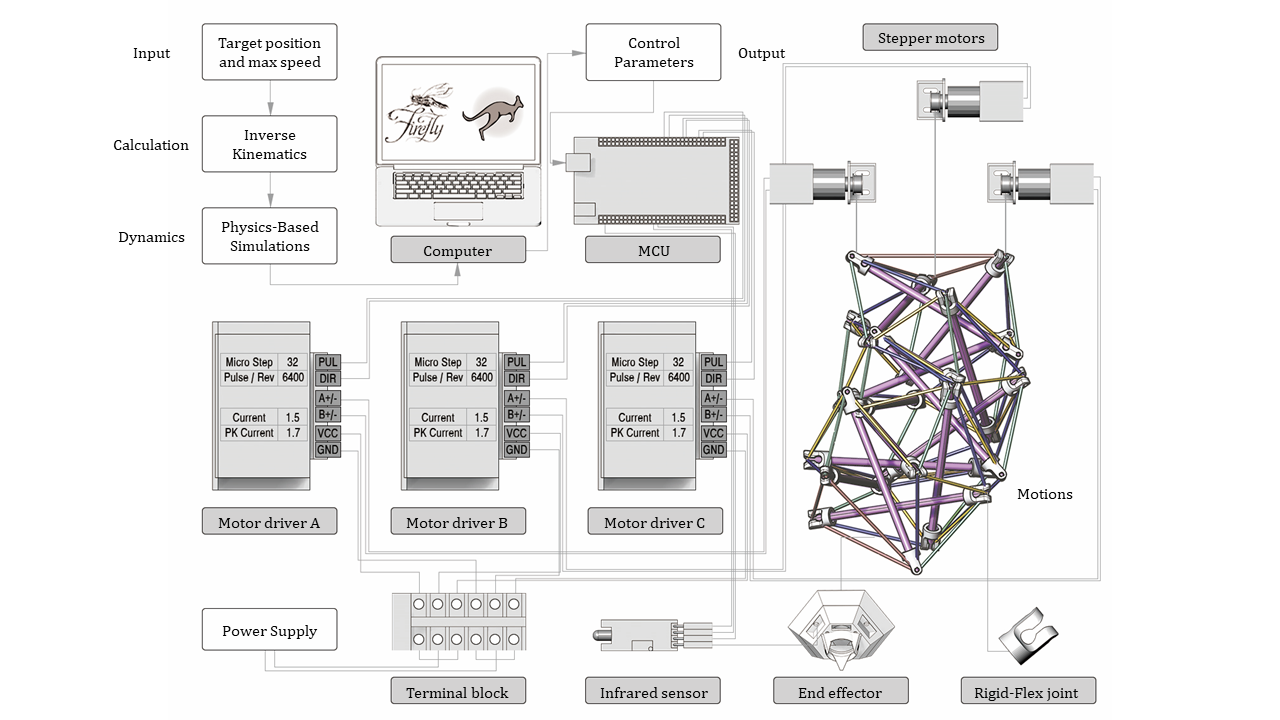}
		\caption{Mechatronic control system.}\label{fig_mechatronic}
	\end{center}
\end{figure}

\begin{figure}[H]
	\begin{center}
		\includegraphics[width=.9\textwidth]{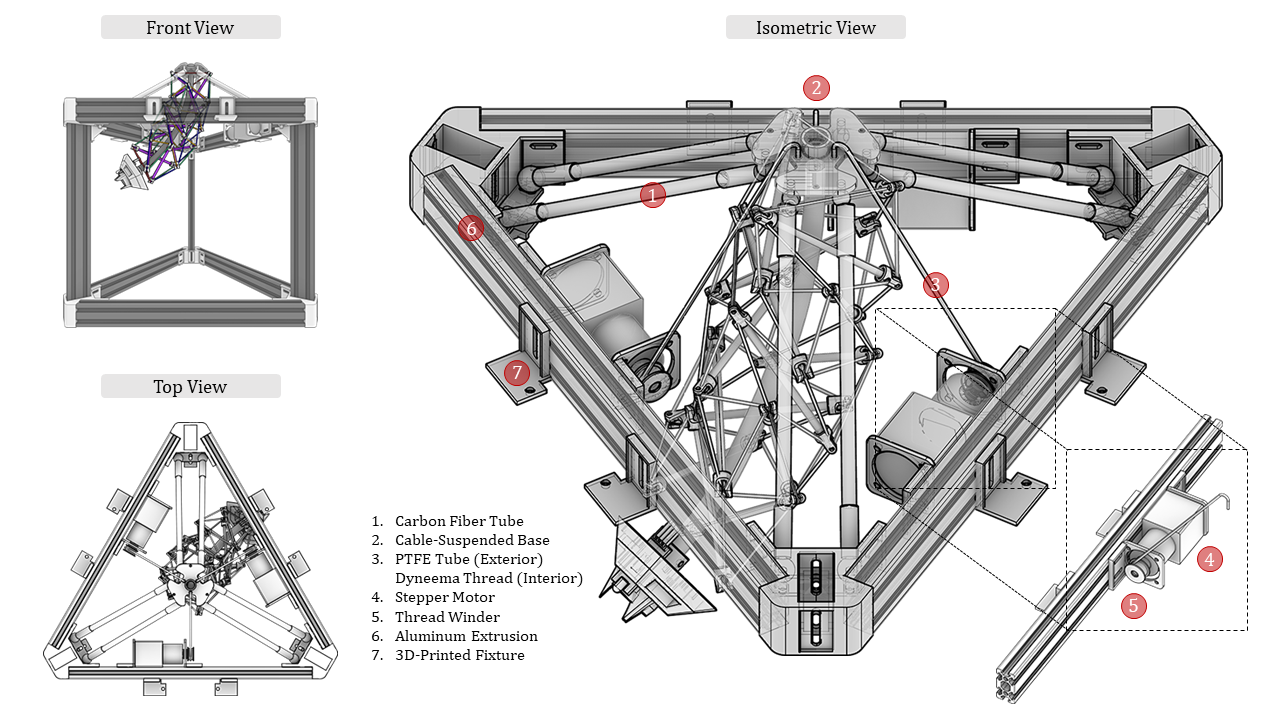}
		\caption{Hardware setup.}\label{fig_hardwareSetup}
	\end{center}
\end{figure}

As demonstrated in Figure~\ref{fig_mechatronic}, the research presents an integrated mechatronic system, including parametric control, real-time simulations, and mechanical actuation, to employ the tensegrity-based compliant mechanism. The robot is controlled by a stepper motor set linked to the computational tools, ensuring distributed cooperative control. Using inverse kinematics and delta robotics, the software platform can calculate the control parameters of each motor to achieve the desired position or configuration. The software toolkit, based on Kangaroo and Firefly in Grasshopper, can conduct real-time simulations according to these control parameters and thus establish relationships between current robot configurations and corresponding data from the actuator.
\par To enhance the trajectory tracking capabilities and other related performance, the entire framework employs a closed-loop control strategy, connecting controller to an infrared sensor. By entering the thermal property of the target, the proposed robot can track the objective and moves adaptively to achieve the trajectory by using the relationship data collected previously. This approach optimizes dexterous mobility and inherent compliance of the tensegrity-based robot body. In addition, equipped with various sensors on the end effector, the robot can be triggered by diverse types of data to fit in its current environment.

\subsection{Hardware Setup}
\begin{figure}[H]
    \centering
    \begin{minipage}[t]{.49\textwidth}
        \centering
        \includegraphics[width=.75\textwidth]{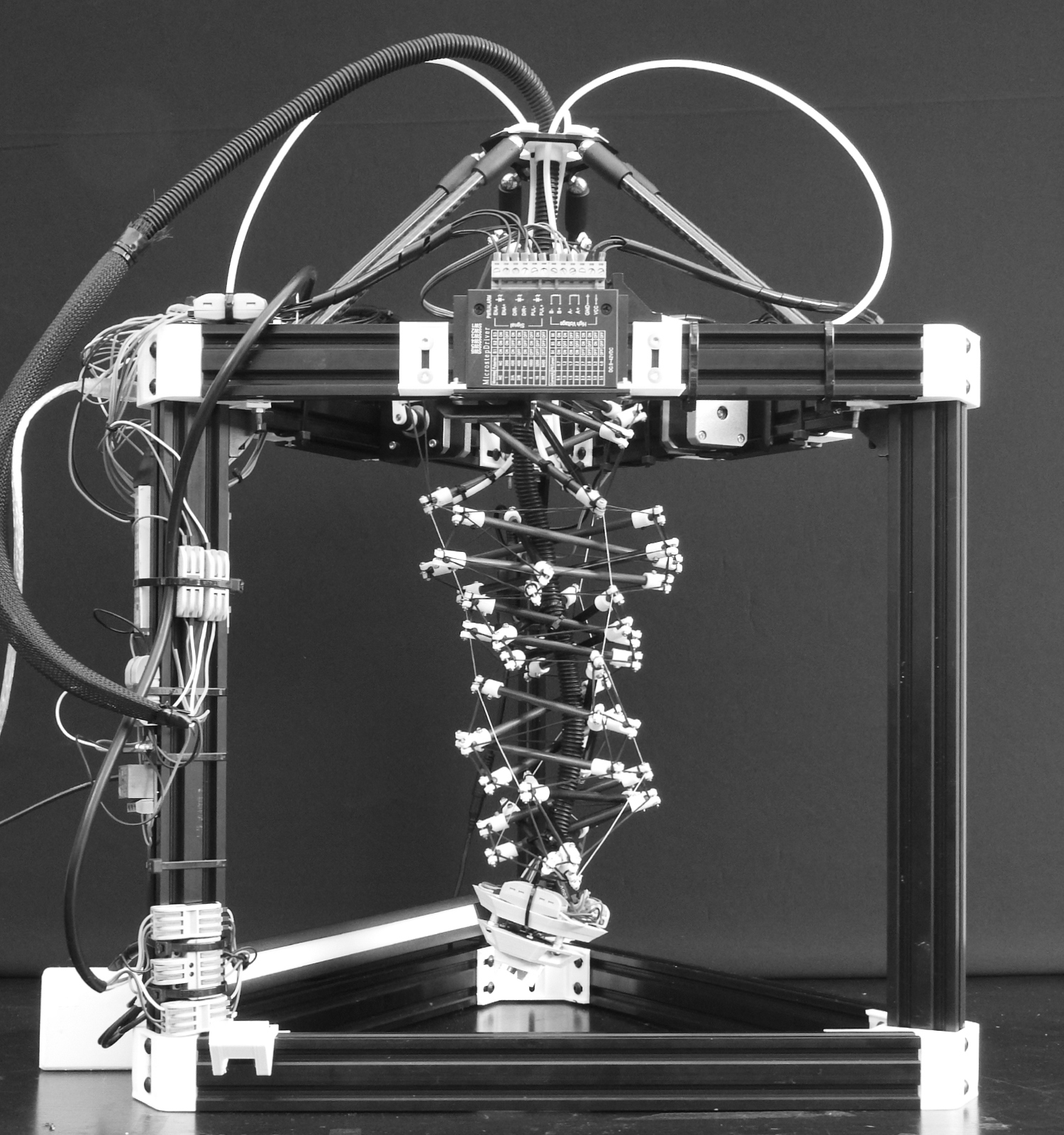}
        \caption{Mechatronic control system.}\label{fig_hardware}
    \end{minipage}
    \hfill
    \begin{minipage}[t]{.49\textwidth}
        \centering
        \includegraphics[width=.75\textwidth]{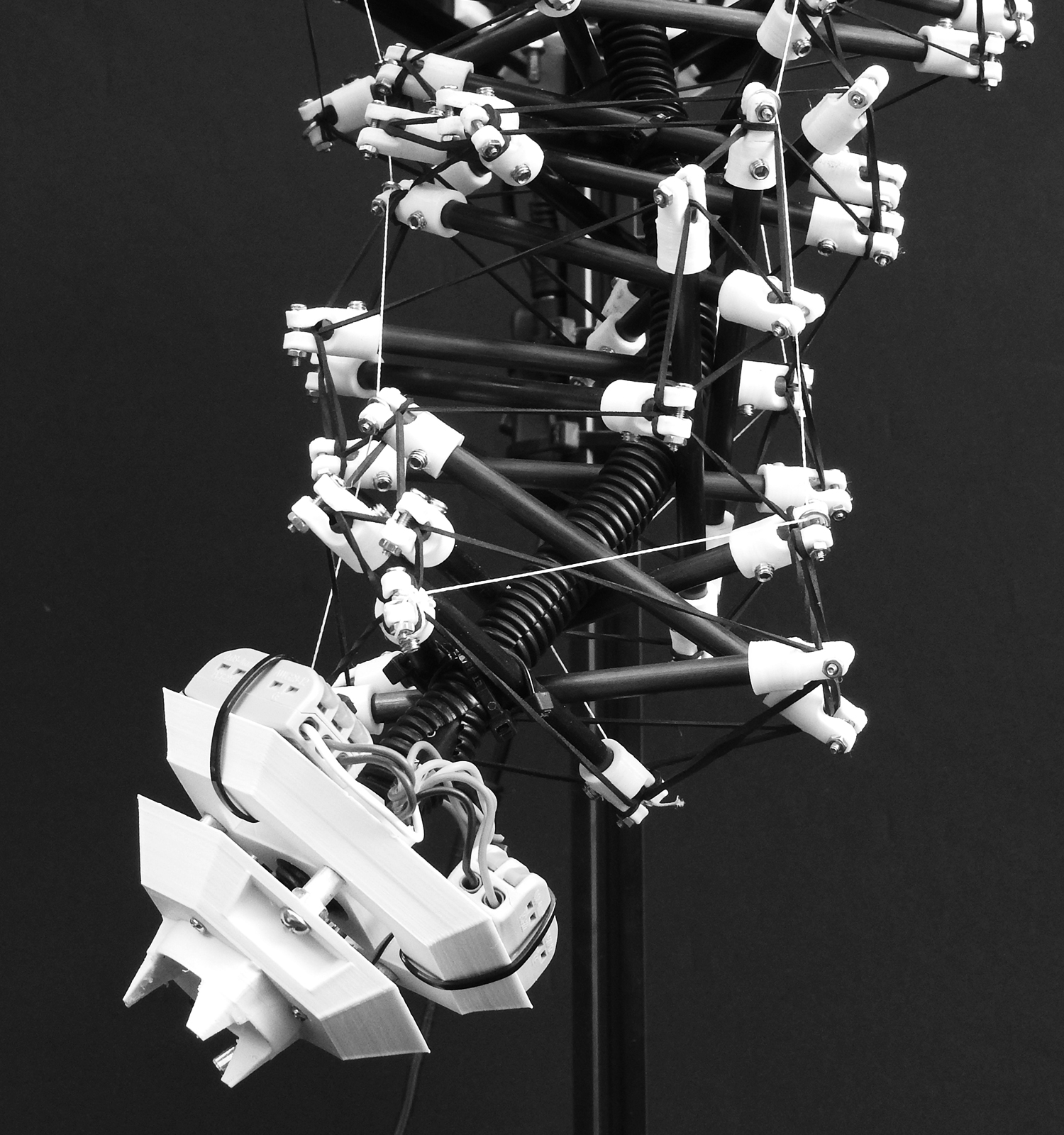}
        \caption{Robot implementation.}\label{fig_endeffector}
    \end{minipage}
\end{figure}

The robot is applied to a triangular platform, as shown in Figure~\ref{fig_hardwareSetup} and~\ref{fig_hardware}, which allows the robot to be precisely controlled using the kinematics solutions (inverse and forward kinematics) of delta robotics. This reduces the complexity of robot programming and ensures the robot can be manipulated intuitively by users. We use inverse kinematics to convert physics simulations into control parameters, and send these data to stepper motors through a microcontroller unit (Arduino). As the motors rotate to desired angle, the tendon actuators are applied to specific pulling forces, resulting length variations and robotic movements. The entire system allows us to synchronize the computational and physical model with control efficiency and accuracy.

\subsection{Material}
The materials used to build up the robot are presented as Table~\ref{table_material}.

\begin{center}
    \begin{longtable}{l
        p{2.5cm}
        p{3cm}
        p{3.5cm}
        p{3cm}}
        \caption{Performance metrics of different learning models} \\
        \toprule
        \textbf{} & \textbf{Cable} & \textbf{Strut} & \textbf{Spine} & \textbf{Actuator}\\
        \midrule
        \endfirsthead
        
        \caption[]{(continued)} \\
        \toprule
		\textbf{} & \textbf{Cable} & \textbf{Strut} & \textbf{Spine} & \textbf{Actuator}\\
        \midrule
        \endhead
        
        \midrule
        \multicolumn{5}{r}{\textit{continued on next page}} \\
        \endfoot
        
        \bottomrule
        \endlastfoot
        
        \textbf{Material} & Rubber Thread & Carbon Fiber Tube & PVC Corrugated Pipe & Dyneema Thread \\
		\label{table_material}
		\vspace{-.5cm}
        
    \end{longtable}
\end{center}

\vspace{-.4cm}

\subsection{End Effector}
The robot is equipped with a mechanical gripper at the end to perform proposed tasks, as shown in Figure~\ref{fig_endeffector}. Moreover, an infrared sensor is embedded in the end effector, allowing the spine robot to receive the distance and thermal data from the current environment.

\subsection{Configurations to Movements}

\begin{figure}[H]
	\begin{center}
		\includegraphics[width=.8\textwidth]{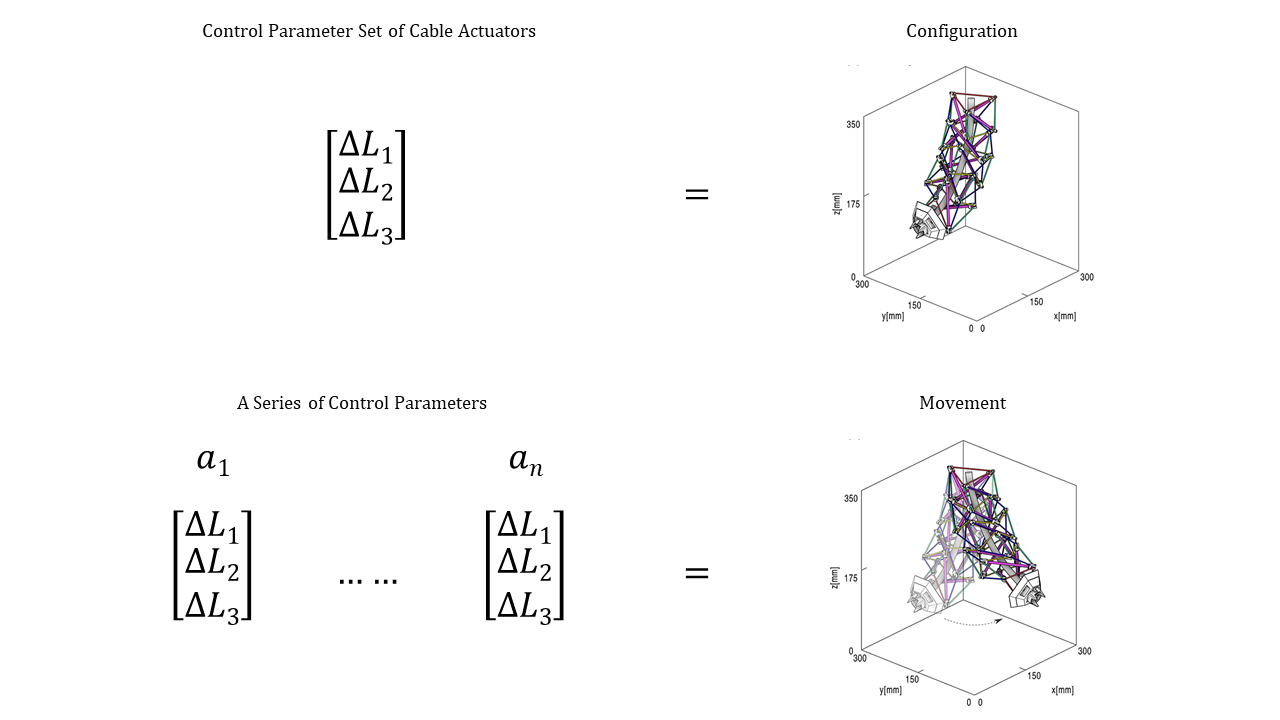}
		\caption{Control method.}\label{fig_controlParameter}
	\end{center}
\end{figure}

\begin{figure}[H]
	\begin{center}
		\includegraphics[width=.8\textwidth]{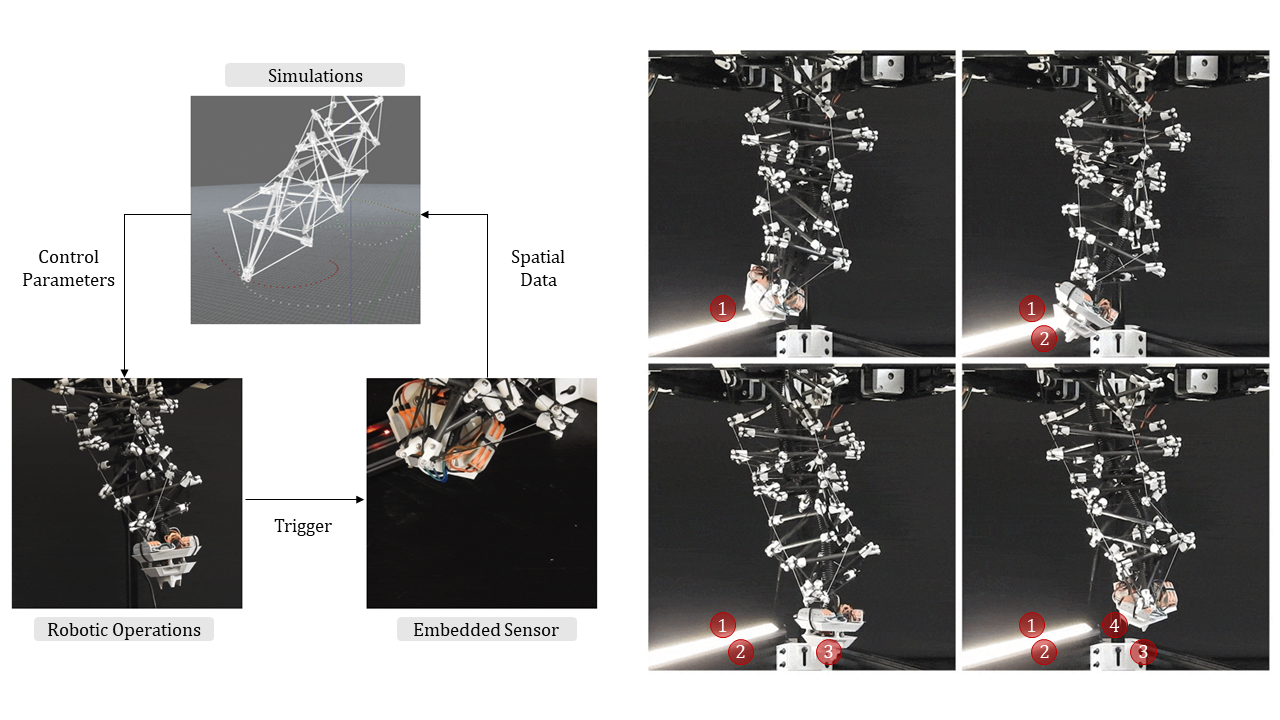}
		\caption{Left: Proposed framework. Right: Actual performance (1-4: target trajectory).}\label{fig_manipulation}
	\end{center}
\end{figure}

Basically, length variations of each tendon actuator ($\Delta$$L_1$, $\Delta$$L_2$, and $\Delta$$L_3$) correspond to a specific configuration (Figure~\ref{fig_controlParameter}). These vairations can be further converted to angular position using the formula for the arc of a circle: $L$ = $\theta$ \texttimes\ $r$, where $\theta$ is the angular position and $r$ is the radius of the thread winder. Thus, the corresponding length varations of the desired configuration can be recognized as a particular control parameter set ($\theta_1$, $\theta_2$, and $\theta_3$). A collection of parameters can eventually be converted into a continuous movement, which is the way we manipulate the robot structure in both computational and physical environments. The framework (Figure~\ref{fig_manipulation}) presents the mothod to validate the performance and to optimize the control system. First, a few sets of control parameters is extracted from simulations and then transferred to the mechatronic implementation. As the robot navigates its environment, the sensors can collect spatial data, which is relayed back to the software for real-time monitoring of the robot's actual position. This end-to-end mapping process from desired to actual trajectories allows for further adjustments, reducing deviations in real-world robotic operations (Figure~\ref{fig_manipulation}).

\section{Results and Discussions}
\subsection{Kinematics}

\begin{figure}[H]
	\begin{center}
		\includegraphics[width=\textwidth]{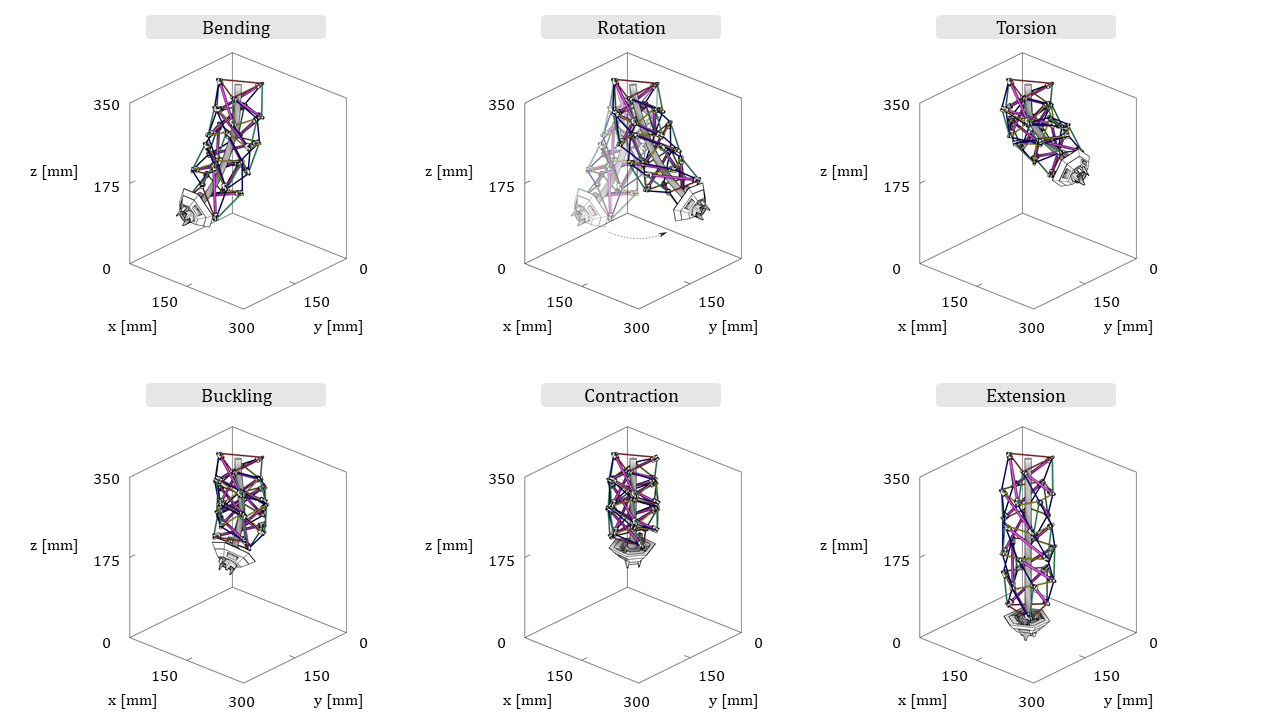}
		\caption{Kinematic simulations.}\label{fig_motion}
	\end{center}
\end{figure}

\begin{figure}[H]
	\begin{center}
		\includegraphics[width=.8\textwidth]{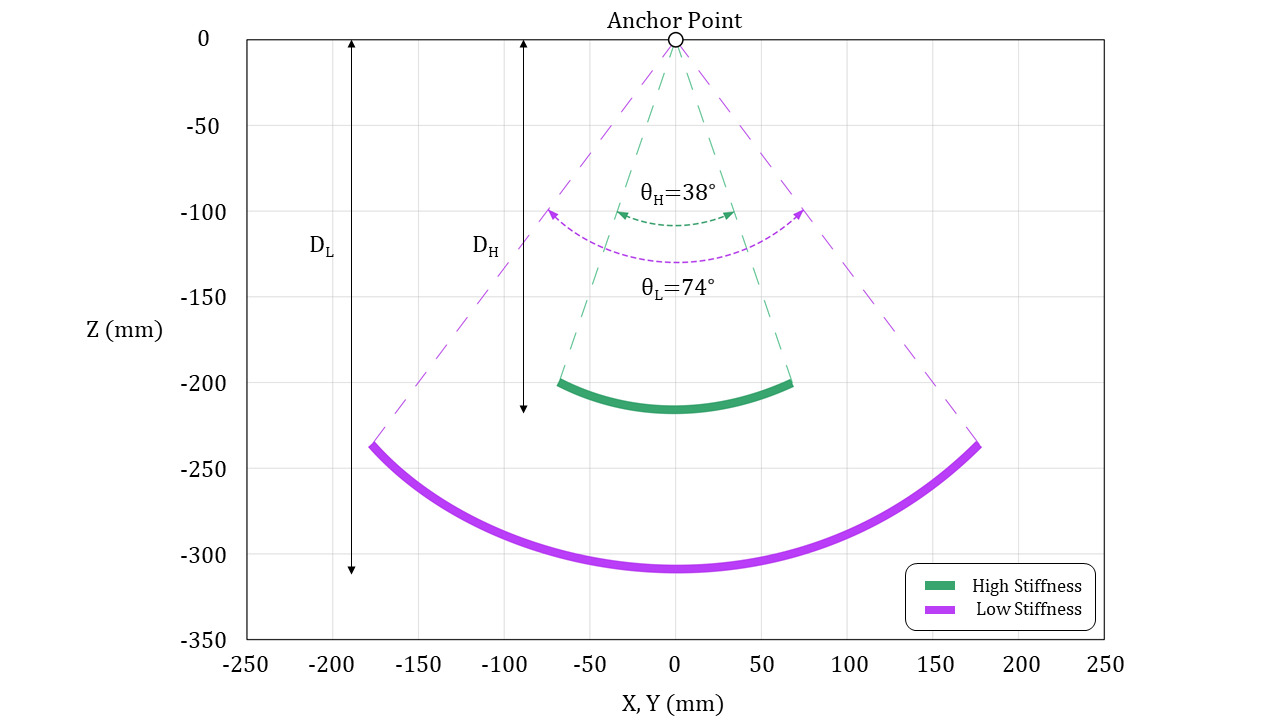}
		\caption{Actual range of motions.}\label{fig_rom}
	\end{center}
\end{figure}

In this research, the mobility of the proposed robot has been explored by leveraging the compliant mechanism. Based on the experimental results, it is validated that the rigid-flex coupling design enhances robotic adaptability. The tensegrity-based robot displays high compliance in degrees-of-freedom, enabling a wide range of motions (ROM).

\par The research has developed a programmable tendon-driven technique, enabling biomimetic locomotion. The self-equilibrating ability allows diverse configuration by applying differential tension force on the actuator. Each set of the control parameters corresponds to a specific configuration, including multidirectional (bending, rotation, and torsion) and unidirectional (buckling, contraction and extension) movements (Figure~\ref{fig_motion}). Most importantly, the combination of different motions enabled the robot to perform more complex movements, such as rotation combined with torsion, bending plus buckling, contraction with bending, etc.

\subsection{Adaptive Stiffness}
Based on the experimental results (Figure~\ref{fig_rom}), it is verified that the adjustments to stiffness can lead to a wide range of motion capabilities. To begin with, the accessible distance is 220mm ($D_H$) with high stiffness, and 310mm ($D_L$) with low stiffness. In addition, the working radius in high and low stiffness state is, respectively, 70mm ($R_H$) and 175mm ($R_L$). Furthermore, the proposed robot exhibits a reachable angle of 38$\deg$ ($\theta_H$) from the primary axis with high stiffness, whereas the low stiffness state provides a maximum of 74$\deg$ ($\theta_L$).

\begin{figure}[H]
	\begin{center}
		\includegraphics[width=\textwidth]{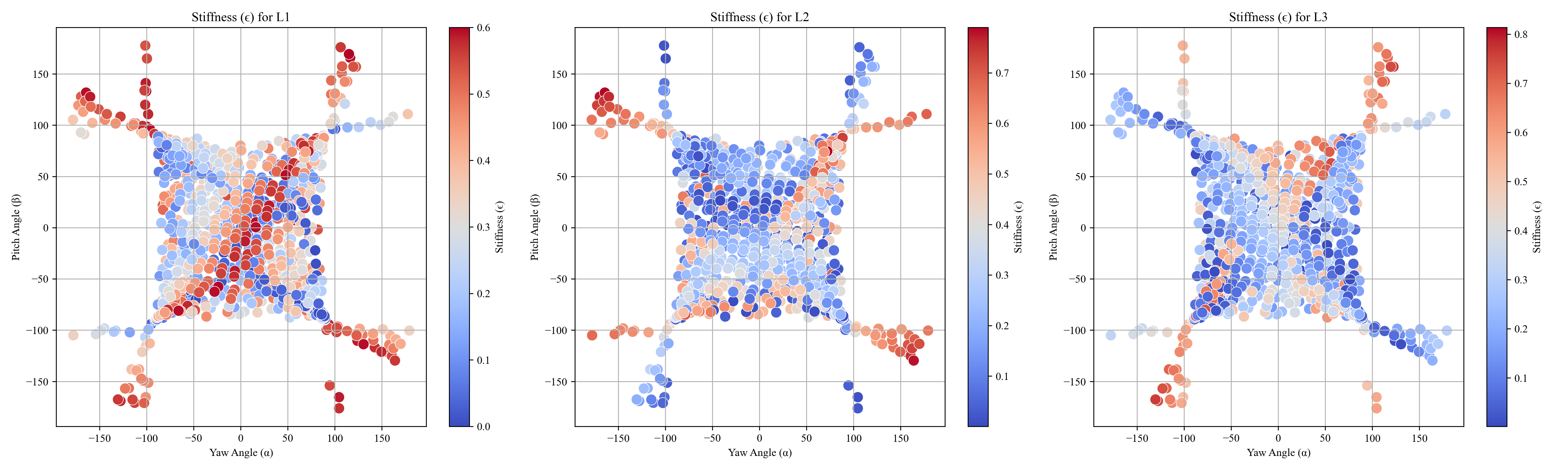}
		\caption{Cable strain by configurations.}\label{fig_stiffness_plots}
	\end{center}
\end{figure}

The research also illustrates the relationship between robot configurations and corresponding cable strain (Figure~\ref{fig_stiffness_plots}). The cable strain ($\epsilon L_1$, $\epsilon L_2$, and $\epsilon L_3$) can be described as the adjustment of cable length divided by the original length ($\epsilon = \Delta L / L$). The configurations are described as ($\alpha$, $\beta$), where $\alpha$ is the yaw angle and $\beta$ is the pitch angle.

\begin{figure}[H]
	\begin{center}
		\includegraphics[width=\textwidth]{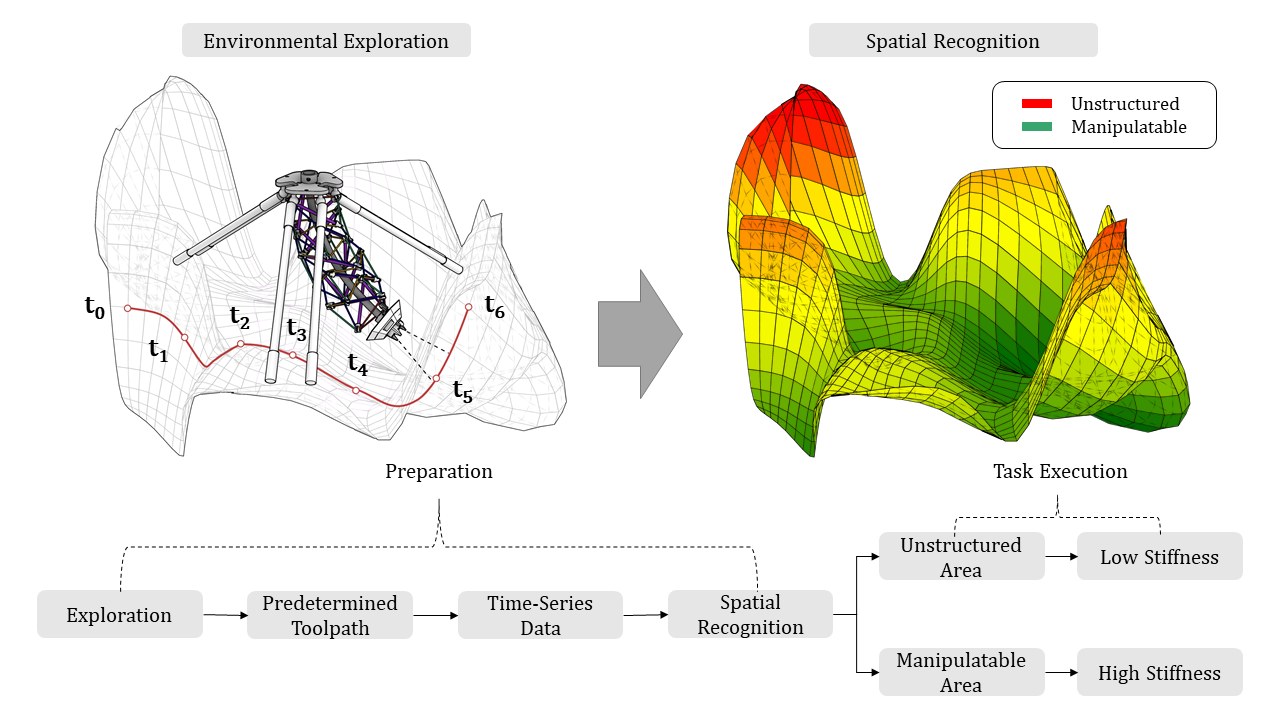}
		\caption{Spatial exploration with adaptive stiffness.}\label{fig_spatialConfig}
	\end{center}
\end{figure}

\subsection{Spatial Exploration}
\begin{figure}[H]
	\begin{center}
		\includegraphics[width=.87\textwidth]{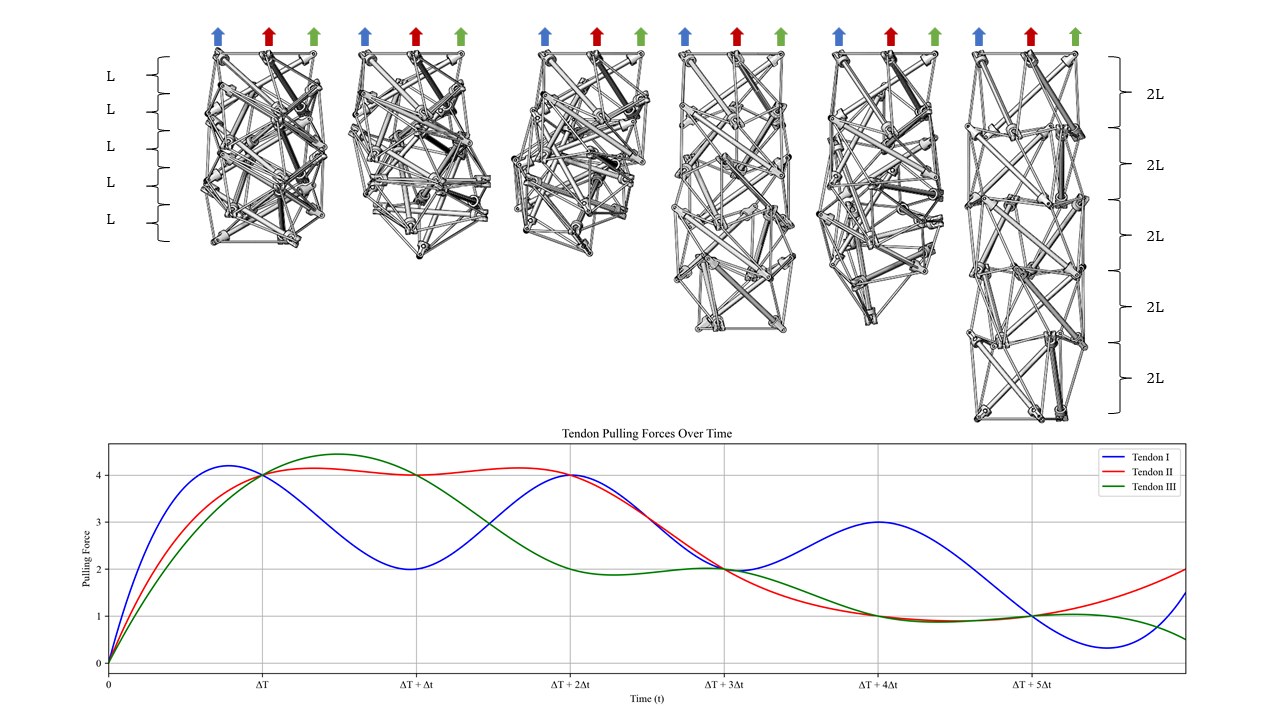}
		\caption{Applied force and movements over time.}\label{fig_forceOverTime}
	\end{center}
\end{figure}

The robot's adaptability enhanced the safety when navigating through unstructured or unfamiliar spaces. As presented in Figure~\ref{fig_spatialConfig}, using the embedded infrared sensor and model-based real-time simulation techniques, the robot can receive time-series data ($t_0$, $t_1$, $t_2$, $\ldots$, $t_n$) including current position, control parameter, stiffness, and distance values between trajectories and the end effector. This enables the computational system to generate a three-dimensional configuration map that distinguishes between unstructured and manipulatable areas. Consequently, the robot gains a better understanding of the current surroundings before task execution and can effectively adjust its stiffness during operation. With each iteration, the robot progressively acquired the capability for biomimetic locomotion. It can conform to obstacles or environmental changes simply by adjusting its stiffness. (Figure~\ref{fig_spatialConfig} and~\ref{fig_forceOverTime}) Significantly, these movements could be achieved with less mechanical complexity.

\begin{figure}[H]
	\begin{center}
		\includegraphics[width=.87\textwidth]{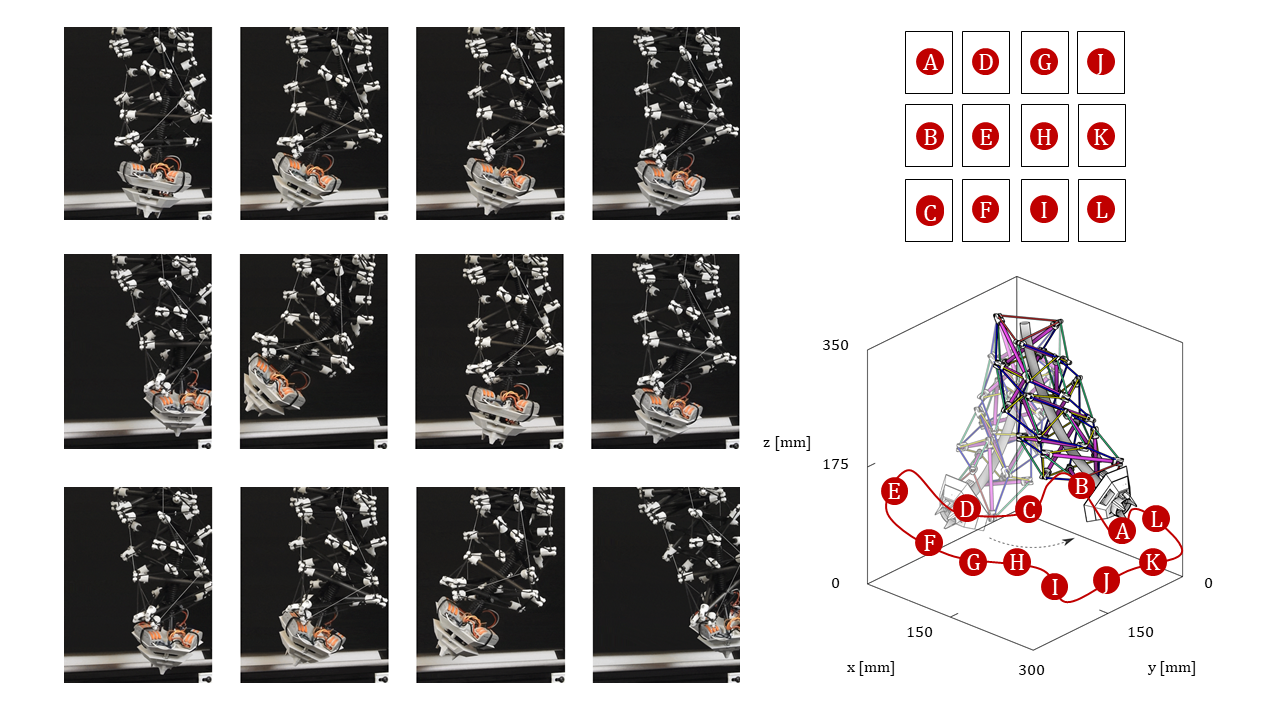}
		\caption{Toolpath.}\label{fig_adaptability}
	\end{center}
\end{figure}

\subsection{Maintenance}

The prestress in the tensional network of tensegrity structures facilitates impact resistance and versatile movement through effective stress distribution. Nevertheless, consideration must be given to the time-dependent prestress loss resulting from cable relaxation. Even slight loss may cause deviation from the simulation results (Figure~\ref{fig_limit_relaxation}).

\subsection{Accuracy}
\begin{figure}[H]
    \centering
    \begin{minipage}[t]{.49\textwidth}
        \centering
        \includegraphics[width=.85\textwidth]{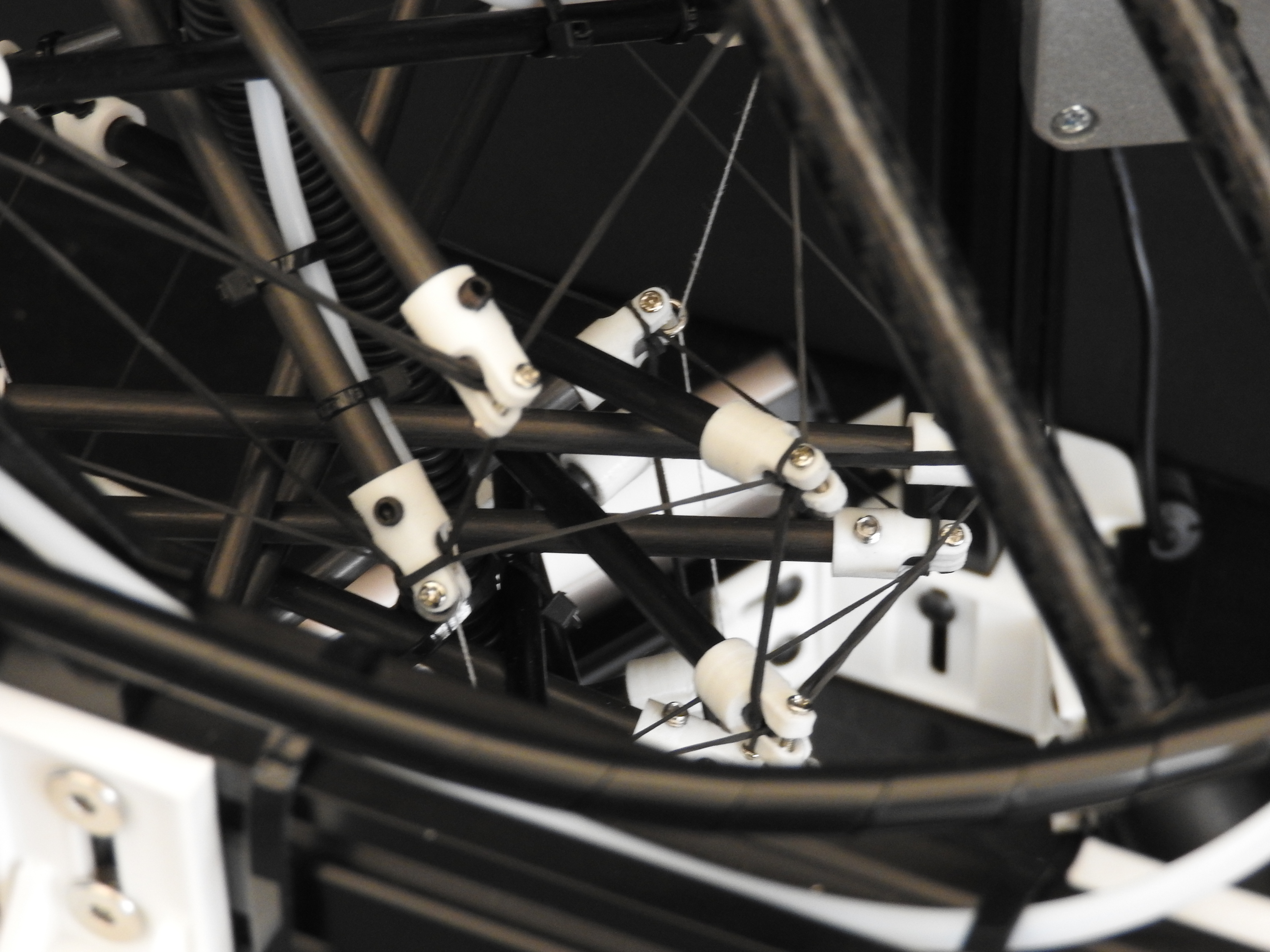}
        \caption{Cable relaxation.}\label{fig_limit_relaxation}
    \end{minipage}
    \hfill
    \begin{minipage}[t]{.49\textwidth}
        \centering
        \includegraphics[width=.85\textwidth]{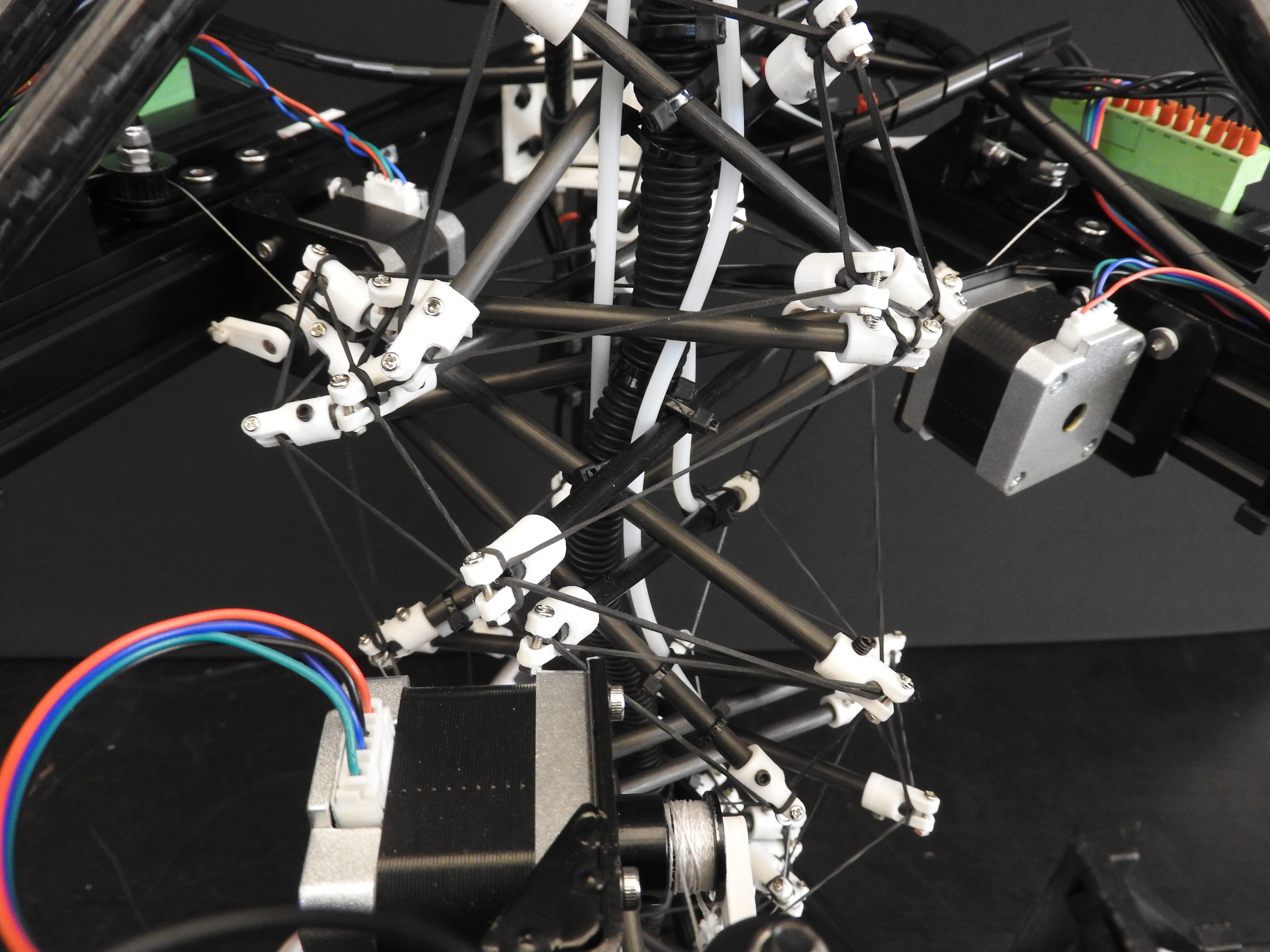}
        \caption{Friction between components.}\label{fig_limit_friction}
    \end{minipage}
\end{figure}

Although the proposed robot validates the effectiveness of the rigid-flex coupling, there are several challenges to be completed before further applications. In fact, the intrinsic mechanical properties of materials may potentially interfere the operation. First, friction between cables and joints may disrupt the movement of the robot, leading to inaccuracies (Figure~\ref{fig_limit_friction}). Second, the risk of cable fracture when exceeding ultimate strength requires attention. This issue may cause the failure of entire tensional network.

\section{Conclusion and Potenital Applications}
To address the constraints encountered by current robots when confronted with external impacts and confined spaces, the research draws inspiration from biological paradigms and explores the implementation of biomimetic robots. By leveraging the musculoskeletal characteristic observed in vertebrate physiology, the research develops a robotic system based on tensegrity structures. This bio-inspired robot embodies a fusion of rigidity and flexibility, actuated by the proposed compliant mechanism. The design not only augments the adaptability to varying environmental conditions but also enhances its kinematic performance. Furthermore, the research introduces a mechatronic control system, integrating computational tools and hardware setup. Utilizing inverse kinematics and physics simulations, the robot can be manipulated to achieve versatile movements and execute diverse missions.
\par The overall robotic performance is evaluated in the research. With variable sets of control parameters, the robot can perform multidirectional and unidirectional movements. Moreover, the passively reconfigurability allows the robot to buffer collisions, attenuating damage from environmental impacts.
\par
\begin{figure}[H]
	\begin{center}
		\includegraphics[width=.85\textwidth]{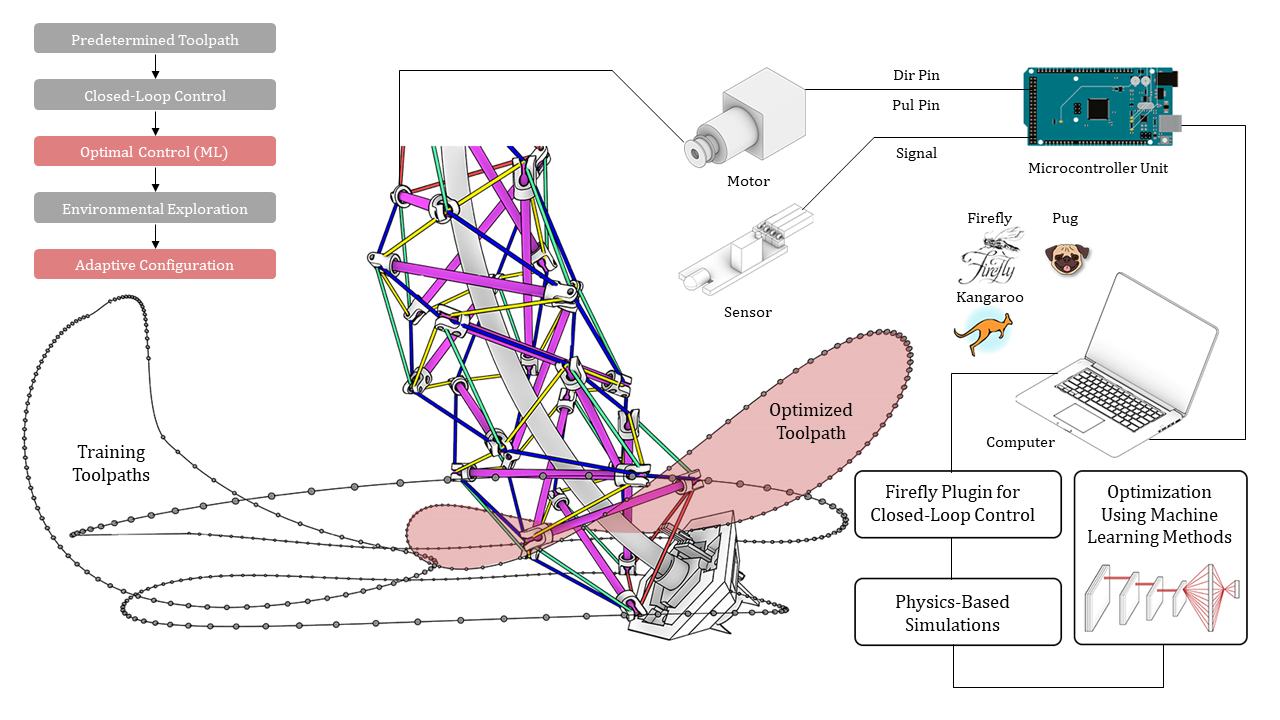}
		\caption{Optimization using machine learning methods.}\label{fig_futureWork}
	\end{center}
\end{figure}

 In terms of future applications, this inherent adaptability enables the robot to complete sophisticated operations in confined environments, such as plumbing shafts, rehabilitation after natural disasters, and underground explorations, demonstrating significant potential for advanced robotic construction. Furthermore, the control strategy may improved by switching from model-based to model-free control (with machine learning methods as presented in Figure~\ref{fig_futureWork}). Through various training processes, the tendon-driven technique can be further optmized.
\par The research also listed several latent issues, such as material deterioration and intrinsic mechanical limitations. However, design iterations with innovative materials, characterized by high tensile strength and durability, still possess a promising trajectory for addressing these challenges.
\par In summation, the research demonstrates a robust foundation for tensegrity-based robots, aiming to explore the robot implementation of biomimicry. The proposed robotic system exhibits noticeable potential to undertake complex tasks in challenging environments, contributing to advancements in digital fabrication.

\section*{Acknowledgements}
This research was supported by XNature: Sensorial Discrete Continuum, funded by the National Science and Technology Council (NSTC 112-2420-H-A49-002). We extend our gratitude to the assistance provided by the Architectural Informatics Lab at the Graduate Institute of Architecture, National Yang Ming Chiao Tung University.

\bibliographystyle{unsrtnat}
\bibliography{references}  

\begin{thebibliography}{11}
\providecommand{\natexlab}[1]{#1}
\providecommand{\url}[1]{\texttt{#1}}
\expandafter\ifx\csname urlstyle\endcsname\relax
  \providecommand{\doi}[1]{doi: #1}\else
  \providecommand{\doi}{doi: \begingroup \urlstyle{rm}\Url}\fi

\bibitem[Trivedi et~al.(2008)Trivedi, Rahn, Kier, and Walker]{trivedi2008soft}
Deepak Trivedi, Christopher~D. Rahn, William~M. Kier, and Ian~D. Walker.
\newblock Soft robotics: Biological inspiration, state of the art, and future
  research.
\newblock \emph{Applied Bionics and Biomechanics}, 5\penalty0 (3):\penalty0
  520417, 2008.
\newblock \doi{10.1080/11762320802557865}.
\newblock URL
  \url{https://onlinelibrary.wiley.com/doi/abs/10.1080/11762320802557865}.

\bibitem[Kobayashi et~al.(2022)Kobayashi, Nabae, Endo, and
  Suzumori]{kobayashi2022soft}
Ryota Kobayashi, Hiroyuki Nabae, Gen Endo, and Koichi Suzumori.
\newblock Soft tensegrity robot driven by thin artificial muscles for the
  exploration of unknown spatial configurations.
\newblock \emph{IEEE Robotics and Automation Letters}, 7\penalty0 (2):\penalty0
  5349--5356, 2022.
\newblock \doi{10.1109/Lra.2022.3153700}.
\newblock URL \url{https://doi.org/10.1109/Lra.2022.3153700}.

\bibitem[Zappetti et~al.(2020)Zappetti, Arandes, Ajanic, and
  Floreano]{zappetti2020variable}
Davide Zappetti, Roc Arandes, Enrico Ajanic, and Dario Floreano.
\newblock Variable-stiffness tensegrity spine.
\newblock \emph{Smart Materials and Structures}, 29\penalty0 (7):\penalty0
  075013, 2020.
\newblock \doi{10.1088/1361-665X/ab87e0}.
\newblock URL \url{https://doi.org/10.1088/1361-665X/ab87e0}.

\bibitem[Liu et~al.(2022)Liu, Bi, Yue, Wu, Yang, and Li]{liu2022review}
Yixiang Liu, Qing Bi, Xiaoming Yue, Jiang Wu, Bin Yang, and Yibin Li.
\newblock A review on tensegrity structures-based robots.
\newblock \emph{Mechanism and Machine Theory}, 168:\penalty0 104571, 2022.
\newblock \doi{10.1016/j.mechmachtheory.2021.104571}.
\newblock URL \url{https://doi.org/10.1016/j.mechmachtheory.2021.104571}.

\bibitem[Lessard et~al.(2016)Lessard, Castro, Asper, Chopra, Baltaxe-Admony,
  Teodorescu, SunSpiral, and Agogino]{lessard2016bio}
Steven Lessard, Dennis Castro, William Asper, Shaurya~Deep Chopra, Leya~Breanna
  Baltaxe-Admony, Mircea Teodorescu, Vytas SunSpiral, and Adrian Agogino.
\newblock A bio-inspired tensegrity manipulator with multi-dof, structurally
  compliant joints.
\newblock In \emph{2016 IEEE/RSJ International Conference on Intelligent Robots
  and Systems (IROS)}, pages 5515--5520. IEEE, 2016.
\newblock \doi{10.1109/IROS.2016.7759811}.
\newblock URL \url{https://doi.org/10.1109/IROS.2016.7759811}.

\bibitem[Motro(2003)]{motro2003tensegrity}
Ren\'e Motro.
\newblock \emph{Tensegrity: structural systems for the future}.
\newblock Elsevier, 2003.

\bibitem[Zhang and Ohsaki(2015)]{zhang2015tensegrity}
Jingyao Zhang and Makoto Ohsaki.
\newblock \emph{Tensegrity structures}, volume~7.
\newblock Springer, 2015.
\newblock URL
  \url{/https://link.springer.com/content/pdf/10.1007/978-4-431-54813-3.pdf}.

\bibitem[Chen and Jiang(2019)]{chen2019swimming}
Bingxing Chen and Hongzhou Jiang.
\newblock Swimming performance of a tensegrity robotic fish.
\newblock \emph{Soft robotics}, 6\penalty0 (4):\penalty0 520--531, 2019.
\newblock \doi{10.1089/soro.2018.0079}.
\newblock URL \url{https://doi.org/10.1089/soro.2018.0079}.
\newblock PMID: 30985267.

\bibitem[Li et~al.(2022)Li, Kim, Park, Choi, Lu, and Peng]{li2022robotic}
Lengxue Li, Sunhong Kim, Junho Park, Youngjin Choi, Qiang Lu, and Dongliang
  Peng.
\newblock Robotic tensegrity structure with a mechanism mimicking human
  shoulder motion.
\newblock \emph{Journal of Mechanisms and Robotics}, 14\penalty0 (2):\penalty0
  025001, 2022.
\newblock \doi{10.1115/1.4052124}.
\newblock URL \url{https://doi.org/10.1115/1.4052124}.

\bibitem[Fasquelle et~al.(2019)Fasquelle, Furet, Chevallereau, and
  Wenger]{fasquelle2019dynamic}
Benjamin Fasquelle, Matthieu Furet, Christine Chevallereau, and Philippe
  Wenger.
\newblock Dynamic modeling and control of a tensegrity manipulator mimicking a
  bird neck.
\newblock In \emph{Advances in Mechanism and Machine Science: Proceedings of
  the 15th IFToMM World Congress on Mechanism and Machine Science 15}, pages
  2087--2097. Springer, 2019.
\newblock \doi{10.1007/978-3-030-20131-9_207}.
\newblock URL \url{https://doi.org/10.1007/978-3-030-20131-9_207}.

\bibitem[Hirose and Mori(2004)]{hirose2004biologically}
Shigeo Hirose and Makoto Mori.
\newblock Biologically inspired snake-like robots.
\newblock In \emph{2004 IEEE International Conference on Robotics and
  Biomimetics}, pages 1--7. IEEE, 2004.
\newblock \doi{10.1109/ROBIO.2004.1521742}.
\newblock URL \url{https://doi.org/10.1109/ROBIO.2004.1521742}.

\end{thebibliography}





\end{document}